\documentclass{article} 
\usepackage{iclr2025_conference,times}

\usepackage{amsmath,amsfonts,bm}

\def\eqref#1{equation~\ref{#1}}

\def\1{\bm{1}}

\DeclareMathAlphabet{\mathsfit}{\encodingdefault}{\sfdefault}{m}{sl}
\SetMathAlphabet{\mathsfit}{bold}{\encodingdefault}{\sfdefault}{bx}{n}

\def\gG{{\mathcal{G}}}

\newcommand{\E}{\mathbb{E}}

\newcommand{\R}{\mathbb{R}}

\usepackage{natbib}
\usepackage{hyperref}
\usepackage{url}
\usepackage{booktabs}
\usepackage{graphicx}
\usepackage{xcolor}
\usepackage[normalem]{ulem}
\hypersetup{colorlinks,citecolor={gray},urlcolor={gray}} 
\title{Clifford Group Equivariant Diffusion Models for 3D Molecular Generation}

\author{\centerline{Cong Liu$^{*}$, Sharvaree Vadgama$^{*}$,} \\ \centerline{\textbf{David Ruhe, Erik Bekkers, Patrick Forr\'e}}\\
\centerline{AMLab, University of Amsterdam, The Netherlands}\\
\textit{\hspace{10mm}${}^*$Equal contribution.}}

\iclrfinalcopy 
\begin{document}

\maketitle
\newcommand{\Euc}{\mathrm{E}}
\newcommand{\SEuc}{\mathrm{SE}}
\newcommand{\Ort}{\mathrm{O}}
\newcommand{\SOrt}{\mathrm{SO}}
\newcommand{\Cl}{\mathrm{Cl}}
\newcommand{\cin}{{c_\text{in}}}
\newcommand{\cout}{{c_\text{out}}}
\newcommand{\vbf}{{\mathbf{v}}}
\newcommand{\wbf}{{\mathbf{w}}}
\newcommand{\fbf}{{\mathbf{f}}}
\newcommand{\phibm}{{\bm{\phi}}}
\newcommand{\psibm}{{\bm{\psi}}}
\newcommand{\bfrak}{{\mathfrak{b}}}
\newcommand{\kbf}{{\mathbf{k}}}
\newcommand{\qbf}{{\mathbf{q}}}
\begin{abstract}
This paper explores leveraging the Clifford algebra's expressive power for $\E(n)$-equivariant diffusion models.
We utilize the geometric products between Clifford multivectors and the rich geometric information encoded in Clifford subspaces in \emph{Clifford Diffusion Models} (CDMs). We extend the diffusion process beyond just Clifford one-vectors to incorporate all higher-grade multivector subspaces. The data is embedded in grade-$k$ subspaces, allowing us to apply latent diffusion across complete multivectors. This enables CDMs to capture the joint distribution across different subspaces of the algebra, incorporating richer geometric information through higher-order features.
We provide empirical results for unconditional molecular generation on the QM9 dataset, showing that CDMs provide a promising avenue for generative modeling.

\end{abstract} 

\section{Introduction}

Deep generative models have revolutionized molecular science, enabling significant progress in molecular design and drug discovery (\cite{Abramson2024, bosese, yim2023se, watson2023rf}). Learning the structure of molecules, these AI models help accelerate drug discovery by replacing costly lab experiments and help streamline the process of designing new drugs or proteins. 

Since molecules exist in 3-dimensional space, thus the group of 3D symmetries $\Euc(3)$, including translations, rotations, and reflections, determines how they transform. To ensure physical validity, generative models for molecular design must respect these symmetries, so that a molecule and any of its symmetric transformations are equally likely according to the learned distribution $p_\theta$, i.e., $p_\theta$ is invariant to $\Euc(3)$ (\cite{xu2022geodiff}). Most recent work addresses this requirement using generative models with denoising neural networks composed of layers that are equivariant to the orthogonal group $\Ort(3)$, generated by rotations and reflections, e.g., equivariant graph neural network (EGNN) layers (\cite{satorras2021n}), which enable the $\Ort(3)$ invariance of the learned distributions (\cite{garcia2021n, hoogeboom2022equivariant, song2024equivariant}). 
However, these methods typically focus only on scalar and scaled Euclidean vector representations, limiting their ability to capture richer geometric structures inherent in molecular systems. 

Clifford Group Equivariant Neural Networks (CGENNs) (\cite{ruhe2023clifford}) establish neural networks operating on the Clifford algebra's rich geometric representations (called \emph{multivectors}) while maintaining $\Ort(n)$- or $\SOrt(n)$-equivariance (\cite{liu2024clifford, brandstetter2022geometric, brehmer2023geometric, zhdanov2024cliffordsteerable, spinner2024lorentz, ruhe2023geometric}). To improve their efficiency, lightweight variants of CGENNs have been introduced (\cite{liu2024multivector}), which make them suitable for large-scale applications.

In this work, we extend diffusion processes into grade-$k$ subspaces, moving beyond conventional vector representations. 
By harnessing multivector structures, our model captures diverse higher-order geometric features of molecular systems by learning on Clifford subspaces. This design enables parallel latent diffusion across distinct Clifford subspaces, facilitating richer symmetry encoding. We demonstrate the efficacy of our approach in unconditional molecular generation tasks on the QM9 dataset, achieving favorable performance compared to existing methods.

\paragraph{Related Work}
Equivariant Diffusion Models (EDMs) \citep{hoogeboom2022equivariant} leverage $\Euc(3)$ symmetries using EGNN as backbone \citep{satorras2021n} and denoising diffusion models \citep{ho2020denoising} to unconditionally generate molecules, atom positions as well as atom types by treating both as continuous variables. Prior to these, works like \citep{gebauer2019symmetry, simonovsky2018graphvae, simm2021symmetryaware} have shown that incorporating symmetries helps generalization for molecular generation. \cite{bekkers2024fast} and \cite{vadgama2025utilityequivariancesymmetrybreaking} use $\SEuc(3)$ equivariant graph backbone and use preconditioned diffusion models \citep{karras2022elucidatingdesignspacediffusionbased} to improve sampling efficiency. GeoLDMs \citep{xu2023geometric} perform latent diffusion with E(3) symmetries in the latent space.
Recent works have focused on a unified representation of discrete and continuous features. \cite{vignac2023midi} generates both molecular graphs and 3D atomic arrangements, while in joint diffusion 2D-3D model, JODO \citep{huang2023learningjoint2d} and flow matching model for generation, FLowMol \citep{dunn2024mixed}, produce complete molecules with atom types, charges, bonds, and 3D coordinates. For the scope of this work, we focus only on using atom positions and atom types as continuous variables and only compare with methods that do so for unconditional generation of molecules. 

\begin{figure}[t]
\begin{center}
\includegraphics[width=0.8\textwidth]{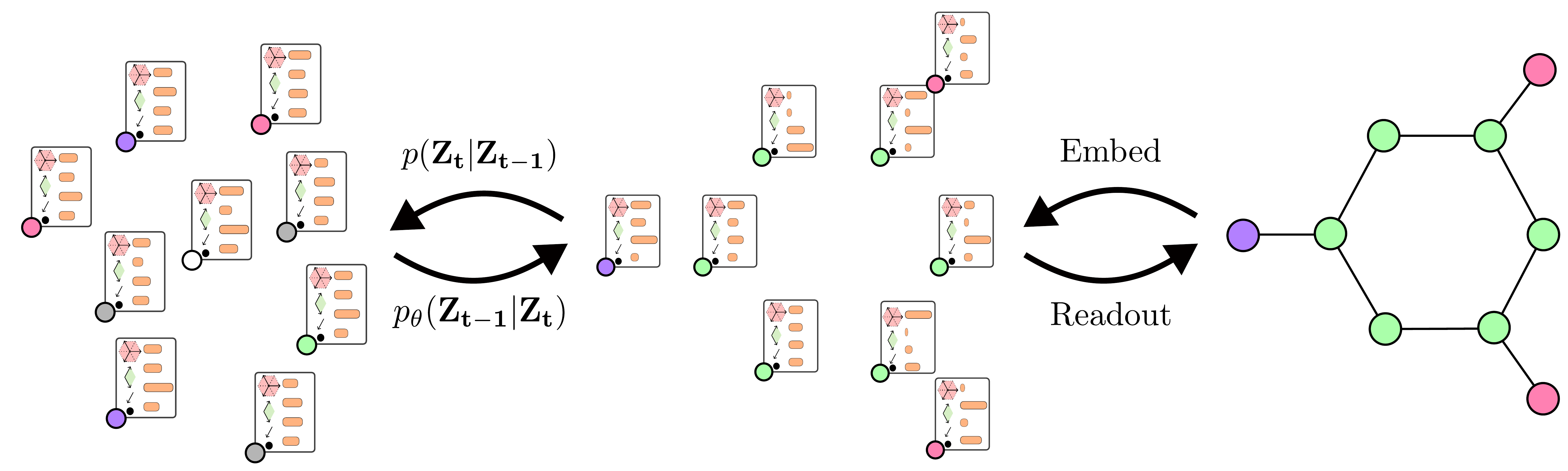}
\end{center}
\caption{During sampling of Clifford all-grade diffusion models, subspace features are initialized from a Gaussian distribution, and CGENNs are used as the denoising model $\phi_\theta$ for each subspace. At time step zero, the molecular structure is read out by projecting a one-vector from Clifford space.}
\end{figure}
\section{Background}

\subsection{$\Euc(3)$ Equivariant Diffusion Models}

$\Euc(3)$ Equivariant Diffusion Models (EDMs) are generative models that learn to transform a simple Gaussian distribution into learned molecular coordinate distribution and atomic type distribution using Denoising Diffusion Probabilistic Models (\cite{ho2020denoising}).Given a data sample $x_1$, during the forward process, the distribution of Gaussian-corrupted sample $x_t$ at timestep \( t \) is defined with noise schedule \( \beta_t \):
\[
q(x_t | x_1) = \mathcal{N}(x_t | \sqrt{\bar{\alpha}_t} x_1, (1 - \bar{\alpha}_t) I),
\]
where \( \bar{\alpha}_t = \prod_{s=1}^t (1 - \beta_s) \). The model predicts the noise \( \epsilon_t \) added to the data during this process and is trained to minimize the denoising loss:
\[
\mathcal{L}_{\text{denoise}} = \mathbb{E}_{t \sim U, x_1 \sim q(x), \epsilon \sim \mathcal{N}(0, I)} \left[ \| \phi_\theta(x_t(x_1), t) - \epsilon \|^2 \right],
\] where $U$ the uniform distribution across predefined time-step range and $q(x_1)$ represents the data distribution, $\phi_\theta$ the neural network that predicts noise. To make the learned distribution $p_\theta$ invariant to $\Euc(3)$, the noise $\epsilon$ and data sample $x_1$ are zero-centered to their mass point. In addition, the network $\phi_\theta$ needs to be equivariant to $\Ort(3)$ transformations (\cite{xu2022geodiff}). Once trained, the model generates samples by reversing the forward process, starting from Gaussian noise and iteratively applying the learned denoising steps.

\subsection{Clifford Group Equivariant Neural Networks}

CGENNs are a class of neural networks that operate on multivectors, rather than traditional scalars or vectors. Built on the mathematical foundation of Clifford (\emph{geometric}) algebra, CGENNs leverage its rich geometric properties to model complex relationships in 3D space, making them well suited for tasks like molecular modeling. The Clifford algebra, denoted as \( \Cl(V, \mathfrak{q}) \), is defined over an \( n \)-dimensional vector space \( V \) paired with a quadratic form \( \mathfrak{q}: V \to \mathbb{R} \). 
In this work, we stick to $V := \mathbb{R}^3$ and use for $\mathfrak{q}$ the standard positive definite Euclidean norm, related to the ordinary Euclidean scalar product. For simplicitly, we write $\Cl(\mathbb{R}^3)$.
In this case, the algebra provides a unified framework to represent scalars, vectors, and higher-dimensional geometric objects, such as bivectors (oriented planes) and trivectors (oriented volumes). The fundamental operation in Clifford algebra is the \emph{geometric product}, which allows vectors to be combined to form higher-grade objects.
A \emph{multivector} is a general element of Clifford algebra, expressed as a direct sum of components from different \emph{grades}, i.e., for $\vbf \in \Cl(\mathbb{R}^3)$:
$
\vbf = a_0 \vbf^{(0)} + a_1 \vbf^{(1)} + a_2 \vbf^{(2)} + a_3 \vbf^{(3)},
$
where \( \vbf^{(m)} \) corresponds to the grade-$m$ component, and $a_i \in \R$ are scalar coefficients. 

\section{Methods}




In this section, we introduce two approaches for extending diffusion models to the Clifford k-grade spaces. First, we demonstrate how diffusion models can be applied to 1-grade subspaces. 
Second, inspired by \cite{xu2023geometric}, we describe how to perform diffusion across all k-grade subspaces of multivectors in a latent embedding, allowing the model to learn and sample the joint distribution of all grades of multivectors in Clifford space.


\subsection{Clifford Group Equivariant One-Vector Diffusion Models}
Consider a molecular graph \( \gG = \{\bm{X, H}\} \), where \( \bm{X} \) represents the atomic coordinates in \( \mathbb{R}^3 \). In Clifford 1-vector diffusion models, we first embed \( \bm{X} \) into the Clifford algebra \( \Cl(\R^3) \) as a 1-vector, denoted as \( \bm{X^{\text{Cl}}} \). Since the grade-one subspace $\Cl^{(1)}(\R^3)$ of \( \Cl(\R^3) \) is isomorphic to \( \mathbb{R}^3 \), this embedding corresponds to a direct identification: each atomic coordinate \( \bm{x}_i \in \mathbb{R}^3 \) is mapped to a 1-vector in $\Cl^{(1)}(\R^3)$, while all higher-grade components (e.g., scalars, bivectors, and trivectors) are set to zero. This ensures that the representation remains consistent with the standard Euclidean vector space while leveraging the algebraic structure of the Clifford space\footnote{We refer to Clifford space as any product of Clifford algebras: $\Cl(\R^3)^k$}.

During the forward process, noise is progressively added to the grade-one component of the Clifford space. The distribution of the intermediate sample \( \bm{X_t}^{\text{Cl}} \) at time step \( t \) follows:
\[
    p(\bm{X^{\text{Cl}}_t} \mid \bm{X^{\text{Cl}}}) = \mathcal{N}(\bm{X^{\text{Cl}}_t}; \sqrt{\bar{\alpha}_t}\bm{X^{\text{Cl}}}, (1 - \bar{\alpha}_t)\bm{I}).
\]

For the backward process, we adopt the Clifford-EGNN introduced in \citep{liu2024multivector} as the denoising model \( \phi_\theta \). This model estimates the noise added to the grade-one components at each time step. The overall diffusion framework follows a structure similar to EDMs, where EDMs employ EGNNs (\cite{satorras2021n}) as denoising models. However, in our approach, \( \bm{X} \) is treated as a multivector, allowing Clifford-EGNN to better capture geometric structures during denoising. Other details of the diffusion process on atomic type features closely follow EDMs.

\subsection{Clifford Group Equivariant All-Grade Diffusion Models}
In Clifford One-Vector Diffusion Models, we treat the Cartesian coordinates \(\bm{X}\) as one-vectors in Clifford space, which is equivalent to viewing \(\bm{X}\) as vectors in Euclidean space. Inspired by \cite{xu2023geometric}, we extend this approach to Clifford All-grade Diffusion Models, where the diffusion process occurs across all Clifford subspaces. To facilitate this, we introduce a latent encoder \( \mathcal{E} \), which lifts \(\bm{X}\) into Clifford space and enriches each sample with geometrically informed features. This enables the diffusion process to leverage higher-order geometric information encoded in different grades of the Clifford algebra. By incorporating these latent geometric features, we aim to assess whether such enrichment provides any significant advantage over traditional diffusion models that operate solely in Euclidean space. 

To obtain higher-order features, we employ a Clifford-EGNN as an encoder $\mathcal{E}$. This encoder takes the initial \(\bm{X^{\text{Cl}}}\) as input and outputs latent multivectors \(\bm{Z}\) with all grades filled with meaningful geometric features. To preserve the geometric layout of the data sample, we replace the grade-1 part of \(\bm{Z}\) with \(\bm{X}\). Consequently, the encoder is responsible for producing only the other grades components of the multivector.
During the forward process, we add Gaussian noise to each subspace of the multivector. Therefore, the distribution of \(\bm{Z_t^{(m)}}\) remains Gaussian: 
\[
p\bigl(\bm{Z_t^{(m)}} \mid \bm{Z^{(m)}}\bigr) = 
\mathcal{N}\bigl(\bm{Z_t^{(m)}} ; \sqrt{\bar{\alpha}_t} \bm{Z^{(m)}}, (1 - \bar{\alpha}_t) \bm{I}\bigr),
\]with $m \in \{0, 1, 2, 3\}$. This means that we add noise independently to each of the Clifford subspaces. The joint distribution $p\bigl(\bm{Z_t} \mid \bm{Z}\bigr)$ then is given by:
\[
p\bigl(\bm{Z_t} \mid \bm{Z}\bigr) = \prod_{m=0}^3 p\bigl(\bm{Z_t^{(m)}} \mid \bm{Z^{(m)}}\bigr).
\]
During the backward process, the denoising model \(\phi_\theta\) learns to output and approximation of the noise \(\bm{\epsilon}\) in Clifford space, i.e.\ \(\bm{\epsilon} \in \Cl(\R^3)\). Thus, both the diffusion (forward) and denoising (backward) steps are fully carried out in Clifford space, leveraging the enriched geometric features at multiple grades.

\section{Experiments}



In this section, we present empirical results for unconditional molecular generation on the QM9 dataset \citep{ramakrishnan2014quantum}. 
 QM9 consists of 3D molecular structures, where each atom is annotated with its atomic type and charge. 
 We evaluate our generation samples on \emph{atomic stability} percentage of generated atoms that satisfy the correct valency, \emph{molecular stability} percentage of molecules in which all constituent atoms are stable, \emph{uniqueness} number of unique samples generated, as well as \emph{validity} percentage of valid samples as defined by RdKit.

    
\begin{table*}[h]
    \small
    \centering
    \resizebox{\textwidth}{!}{%
    \begin{tabular}{lcccc}
    \toprule
    Models & Atom Stability ($\%$) & Mol Stability ($\%$) & Validity ($\%$) & Valid \& Unique ($\%$) \\ \midrule
    GDM-AUG & $97.6$ & $71.6$ & $90.4$ & $89.5$ \\
    EDM (\cite{hoogeboom2022equivariant}) & $98.7 \pm 0.1 $  &      $82.0 \pm 0.4$  & $91.9 \pm 0.5$ & $83.3 \pm 0.6$   \\ 
    
    GeoLDM (\cite{xu2023geometric})  & $98.9 \pm 0.1 $  &      $89.4 \pm 0.5$  & $93.8 \pm 0.4$ & $92.7 \pm 0.5$   \\ 
    P$\Theta$NITA (\cite{bekkers2024fast}) & $98.9$ & $87.8$   & - & - \\
    MUDiff$^\dagger$(\cite{hua2024mudiff}) & $98.8\pm0.2$ & $89.9 \pm 1.1$ & $95.3 \pm 1.5$ & $94.4 \pm 0.5$ \\
    END$^\dagger$ (\cite{cornet2024equivariant}) & $98.9\pm0.2$ & $89.1 \pm 0.1$ & $94.8 \pm 1.5$ & $87.8 \pm 0.4$ \\
    
    EquiFM$^\dagger$ (\cite{song2024equivariant}) & $98.9\pm0.1$ & $88.3 \pm 0.3$ & $94.7 \pm 0.4$ & $93.5 \pm 0.3$ \\
    Rapidash$^\dagger$ (\cite{vadgama2025utilityequivariancesymmetrybreaking}) & $\bm{99.38 \pm 0.02}$ & $\bm{92.91 \pm 0.41}$ & $ \bm{98.12 \pm .003} $ & $95.35 \pm .003$\\ 
    \midrule
    
    \textbf{CDM (one-vector)} & $98.9 \pm 0.0$  &    $89.6 \pm 0.2 $ & $96.0 \pm 0.3$  & $95.8 \pm 0.3$    \\ 
    \textbf{CDM (all-grade) } & $99.0 \pm 0.2 $  &    $89.7 \pm 1.4 $ & $96.4 \pm 1.0$  & $\bm{96.3 \pm 1.0}$    \\ \midrule
    Data  & $99.00$ &  $95.20$ & $97.8$ & $97.8$ \\ \bottomrule
    
    \end{tabular}
    }
    \caption{Results for unconditional generation task on QM9 dataset. $^\dagger$ represents works that have a different generative model than the rest. }
    \label{tab:qm9}
\end{table*}

We compare our proposed models, CDMs\footnote{To ensure a fair comparison, our CDMs have equal or less parameter counts compared to baseline models.} with Graph Diffusion Models without symmetries (GDM -AUG),  E(3) equivariant diffusion models EDMs, \citep{hoogeboom2022equivariant}, latent diffusion models, GeoLDMs \citep{xu2023geometric}, MuDiff \citep{hua2024mudiff}, END \citep{cornet2024equivariant}, EquiFM \citep{song2024equivariant}, and Rapidash \citep{vadgama2025utilityequivariancesymmetrybreaking}.
During evaluation, 10000 samples are generated from each models to evaluate the standard metrics. From Table \ref{tab:qm9}, we can see that our CDMs perform competitively across all evaluated metrics compared to baseline models, which validates our models' effectiveness. CDMs with all-grade diffusion in general generate molecules with higher quality. This result indicates strong potential of applying generative models on other grades in Clifford algebra. All CDM results are calculated based on three runs with different seeds. 

\section{Conclusion}
We introduce Clifford Group Equivariant Diffusion Models (CDMs), a diffusion framework operating in Clifford grade-$k$ (sub)spaces for unconditional molecular generation. Among the two variants, Clifford one-vector diffusion effectively captures molecular geometry while maintaining a direct correspondence with Euclidean space. To compare with \cite{xu2023geometric}, we explored Clifford all-grade diffusion, where a geometric encoder maps molecular structures into a latent Clifford representation, initializing latent Clifford features -- scalars, bivectors, and trivectors before diffusion. While its effectiveness compared to Clifford one-vector diffusion remains under evaluation, we believe encoding higher-order geometric structures could enhance generative modeling in more complex scenarios.




\section{Acknowledgements}
C. Liu is supported by Health-Holland, Top Sector Life Sciences \& Health (LSH-TKI), project number LSHM22023, which realizes a public-private partnership (PPP) between the University of Amsterdam and Janssen Vaccines and Prevention B.V. S. Vadgama is supported by the Hybrid Intelligence Center, a 10-year program funded by the Dutch Ministry of Education, Culture and Science through the Netherlands Organisation for Scientific Research (NWO). We thank SURF (www.surf.nl) for the support in using the National Supercomputer Snellius.

\bibliography{iclr2025_conference}

\begin{thebibliography}{31}
\providecommand{\natexlab}[1]{#1}
\providecommand{\url}[1]{\texttt{#1}}
\expandafter\ifx\csname urlstyle\endcsname\relax
  \providecommand{\doi}[1]{doi: #1}\else
  \providecommand{\doi}{doi: \begingroup \urlstyle{rm}\Url}\fi

\bibitem[Abramson et~al.(2024)Abramson, Adler, Dunger, Evans, Green, Pritzel,
  Ronneberger, Willmore, Ballard, Bambrick, Bodenstein, Evans, Hung, O’Neill,
  Reiman, Tunyasuvunakool, Wu, Žemgulytė, Arvaniti, Beattie, Bertolli,
  Bridgland, Cherepanov, Congreve, Cowen-Rivers, Cowie, Figurnov, Fuchs,
  Gladman, Jain, Khan, Low, Perlin, Potapenko, Savy, Singh, Stecula,
  Thillaisundaram, Tong, Yakneen, Zhong, Zielinski, Žídek, Bapst, Kohli,
  Jaderberg, Hassabis, and Jumper]{Abramson2024}
Josh Abramson, Jonas Adler, Jack Dunger, Richard Evans, Tim Green, Alexander
  Pritzel, Olaf Ronneberger, Lindsay Willmore, Andrew~J. Ballard, Joshua
  Bambrick, Sebastian~W. Bodenstein, David~A. Evans, Chia-Chun Hung, Michael
  O’Neill, David Reiman, Kathryn Tunyasuvunakool, Zachary Wu, Akvilė
  Žemgulytė, Eirini Arvaniti, Charles Beattie, Ottavia Bertolli, Alex
  Bridgland, Alexey Cherepanov, Miles Congreve, Alexander~I. Cowen-Rivers,
  Andrew Cowie, Michael Figurnov, Fabian~B. Fuchs, Hannah Gladman, Rishub Jain,
  Yousuf~A. Khan, Caroline M.~R. Low, Kuba Perlin, Anna Potapenko, Pascal Savy,
  Sukhdeep Singh, Adrian Stecula, Ashok Thillaisundaram, Catherine Tong, Sergei
  Yakneen, Ellen~D. Zhong, Michal Zielinski, Augustin Žídek, Victor Bapst,
  Pushmeet Kohli, Max Jaderberg, Demis Hassabis, and John~M. Jumper.
\newblock Accurate structure prediction of biomolecular interactions with
  alphafold 3.
\newblock \emph{Nature}, 630\penalty0 (8016), 2024.
\newblock \doi{10.1038/s41586-024-07487-w}.

\bibitem[Bekkers et~al.(2024)Bekkers, Vadgama, Hesselink, der Linden, and
  Romero]{bekkers2024fast}
Erik~J Bekkers, Sharvaree Vadgama, Rob Hesselink, Putri A~Van der Linden, and
  David~W. Romero.
\newblock Fast, expressive se(n) equivariant networks through weight-sharing in
  position-orientation space.
\newblock In \emph{The Twelfth International Conference on Learning
  Representations}, 2024.

\bibitem[Bose et~al.(2024)Bose, Akhound-Sadegh, Huguet, FATRAS, Rector-Brooks,
  Liu, Nica, Korablyov, Bronstein, and Tong]{bosese}
Joey Bose, Tara Akhound-Sadegh, Guillaume Huguet, Kilian FATRAS, Jarrid
  Rector-Brooks, Cheng-Hao Liu, Andrei~Cristian Nica, Maksym Korablyov,
  Michael~M Bronstein, and Alexander Tong.
\newblock Se (3)-stochastic flow matching for protein backbone generation.
\newblock In \emph{The Twelfth International Conference on Learning
  Representations}, 2024.

\bibitem[Brandstetter et~al.(2022)Brandstetter, Hesselink, van~der Pol,
  Bekkers, and Welling]{brandstetter2022geometric}
Johannes Brandstetter, Rob Hesselink, Elise van~der Pol, Erik~J Bekkers, and
  Max Welling.
\newblock {Geometric and Physical Quantities Improve {E}(3) Equivariant Message
  Passing}.
\newblock In \emph{International Conference on Learning Representations}, 2022.

\bibitem[Brehmer et~al.(2023)Brehmer, de~Haan, Behrends, and
  Cohen]{brehmer2023geometric}
Johann Brehmer, Pim de~Haan, S{\"o}nke Behrends, and Taco Cohen.
\newblock {Geometric Algebra Transformer}.
\newblock In \emph{Neural Information Processing Systems}, 2023.

\bibitem[Cornet et~al.(2024)Cornet, Bartosh, Schmidt, and
  Naesseth]{cornet2024equivariant}
Fran{\c{c}}ois Cornet, Grigory Bartosh, Mikkel~N Schmidt, and Christian~A
  Naesseth.
\newblock Equivariant neural diffusion for molecule generation.
\newblock In \emph{38th Conference on Neural Information Processing Systems},
  2024.

\bibitem[Dunn \& Koes(2024)Dunn and Koes]{dunn2024mixed}
Ian Dunn and David~Ryan Koes.
\newblock Mixed continuous and categorical flow matching for 3d de novo
  molecule generation.
\newblock \emph{ArXiv}, 2024.

\bibitem[Garcia~Satorras et~al.(2021)Garcia~Satorras, Hoogeboom, Fuchs, Posner,
  and Welling]{garcia2021n}
Victor Garcia~Satorras, Emiel Hoogeboom, Fabian Fuchs, Ingmar Posner, and Max
  Welling.
\newblock E (n) equivariant normalizing flows.
\newblock \emph{Advances in Neural Information Processing Systems}, 2021.

\bibitem[Gebauer et~al.(2019)Gebauer, Gastegger, and
  Sch{\"u}tt]{gebauer2019symmetry}
Niklas Gebauer, Michael Gastegger, and Kristof Sch{\"u}tt.
\newblock Symmetry-adapted generation of 3d point sets for the targeted
  discovery of molecules.
\newblock \emph{Advances in neural information processing systems}, 2019.

\bibitem[Ho et~al.(2020)Ho, Jain, and Abbeel]{ho2020denoising}
Jonathan Ho, Ajay Jain, and Pieter Abbeel.
\newblock Denoising diffusion probabilistic models.
\newblock \emph{Advances in neural information processing systems}, 2020.

\bibitem[Hoogeboom et~al.(2022)Hoogeboom, Satorras, Vignac, and
  Welling]{hoogeboom2022equivariant}
Emiel Hoogeboom, V{\i}ctor~Garcia Satorras, Cl{\'e}ment Vignac, and Max
  Welling.
\newblock Equivariant diffusion for molecule generation in 3d.
\newblock In \emph{International conference on machine learning}, 2022.

\bibitem[Hua et~al.(2024)Hua, Luan, Xu, Ying, Fu, Ermon, and
  Precup]{hua2024mudiff}
Chenqing Hua, Sitao Luan, Minkai Xu, Zhitao Ying, Jie Fu, Stefano Ermon, and
  Doina Precup.
\newblock Mudiff: Unified diffusion for complete molecule generation.
\newblock In \emph{Learning on Graphs Conference}, pp.\  33--1. PMLR, 2024.

\bibitem[Huang et~al.(2023)Huang, Sun, Du, and Lv]{huang2023learningjoint2d}
Han Huang, Leilei Sun, Bowen Du, and Weifeng Lv.
\newblock Learning joint 2d \& 3d diffusion models for complete molecule
  generation, 2023.

\bibitem[Karras et~al.(2022)Karras, Aittala, Aila, and
  Laine]{karras2022elucidatingdesignspacediffusionbased}
Tero Karras, Miika Aittala, Timo Aila, and Samuli Laine.
\newblock Elucidating the design space of diffusion-based generative models,
  2022.

\bibitem[Liu et~al.(2024{\natexlab{a}})Liu, Ruhe, Eijkelboom, and
  Forr{\'e}]{liu2024clifford}
Cong Liu, David Ruhe, Floor Eijkelboom, and Patrick Forr{\'e}.
\newblock {Clifford Group Equivariant Simplicial Message Passing Networks}.
\newblock In \emph{The Twelfth International Conference on Learning
  Representations}, 2024{\natexlab{a}}.

\bibitem[Liu et~al.(2024{\natexlab{b}})Liu, Ruhe, and
  Forr{\'e}]{liu2024multivector}
Cong Liu, David Ruhe, and Patrick Forr{\'e}.
\newblock Multivector neurons: Better and faster o(n)-equivariant clifford
  {GNN}s.
\newblock In \emph{ICML 2024 Workshop on Geometry-grounded Representation
  Learning and Generative Modeling}, 2024{\natexlab{b}}.

\bibitem[Ramakrishnan et~al.(2014)Ramakrishnan, Dral, Rupp, and
  Von~Lilienfeld]{ramakrishnan2014quantum}
Raghunathan Ramakrishnan, Pavlo~O Dral, Matthias Rupp, and O~Anatole
  Von~Lilienfeld.
\newblock Quantum chemistry structures and properties of 134 kilo molecules.
\newblock \emph{Scientific data}, \penalty0 (1), 2014.

\bibitem[Ruhe et~al.(2023{\natexlab{a}})Ruhe, Brandstetter, and
  Forr\'e]{ruhe2023clifford}
David Ruhe, Johannes Brandstetter, and Patrick Forr\'e.
\newblock {Clifford Group Equivariant Neural Networks}.
\newblock In \emph{Neural Information Processing Systems}, 2023{\natexlab{a}}.

\bibitem[Ruhe et~al.(2023{\natexlab{b}})Ruhe, Gupta, Keninck, Welling, and
  Brandstetter]{ruhe2023geometric}
David Ruhe, Jayesh~K. Gupta, Steven~De Keninck, Max Welling, and Johannes
  Brandstetter.
\newblock {Geometric Clifford Algebra Networks}.
\newblock In \emph{International Conference on Machine Learning},
  2023{\natexlab{b}}.

\bibitem[Satorras et~al.(2021)Satorras, Hoogeboom, and Welling]{satorras2021n}
V{\i}ctor~Garcia Satorras, Emiel Hoogeboom, and Max Welling.
\newblock E (n) equivariant graph neural networks.
\newblock In \emph{International conference on machine learning}, 2021.

\bibitem[Simm et~al.(2021)Simm, Pinsler, Cs{\'a}nyi, and
  Hern{\'a}ndez-Lobato]{simm2021symmetryaware}
Gregor N.~C. Simm, Robert Pinsler, G{\'a}bor Cs{\'a}nyi, and Jos{\'e}~Miguel
  Hern{\'a}ndez-Lobato.
\newblock Symmetry-aware actor-critic for 3d molecular design.
\newblock In \emph{International Conference on Learning Representations}, 2021.

\bibitem[Simonovsky \& Komodakis(2018)Simonovsky and
  Komodakis]{simonovsky2018graphvae}
Martin Simonovsky and Nikos Komodakis.
\newblock Graphvae: Towards generation of small graphs using variational
  autoencoders.
\newblock In \emph{International Conference on Artificial Neural Networks},
  2018.

\bibitem[Song et~al.(2024)Song, Gong, Xu, Cao, Lan, Ermon, Zhou, and
  Ma]{song2024equivariant}
Yuxuan Song, Jingjing Gong, Minkai Xu, Ziyao Cao, Yanyan Lan, Stefano Ermon,
  Hao Zhou, and Wei-Ying Ma.
\newblock Equivariant flow matching with hybrid probability transport for 3d
  molecule generation.
\newblock \emph{Advances in Neural Information Processing Systems}, 36, 2024.

\bibitem[Spinner et~al.(2024)Spinner, Bres{\'o}, de~Haan, Plehn, Thaler, and
  Brehmer]{spinner2024lorentz}
Jonas Spinner, Victor Bres{\'o}, Pim de~Haan, Tilman Plehn, Jesse Thaler, and
  Johann Brehmer.
\newblock {Lorentz-Equivariant Geometric Algebra Transformers for High-Energy
  Physics}.
\newblock 2024.

\bibitem[Vadgama et~al.(2025)Vadgama, Islam, Buracus, Shewmake, and
  Bekkers]{vadgama2025utilityequivariancesymmetrybreaking}
Sharvaree Vadgama, Mohammad~Mohaiminul Islam, Domas Buracus, Christian
  Shewmake, and Erik Bekkers.
\newblock On the utility of equivariance and symmetry breaking in deep learning
  architectures on point clouds, 2025.

\bibitem[Vignac et~al.(2023)Vignac, Osman, Toni, and Frossard]{vignac2023midi}
Clement Vignac, Nagham Osman, Laura Toni, and Pascal Frossard.
\newblock Midi: Mixed graph and 3d denoising diffusion for molecule generation.
\newblock In \emph{Joint European Conference on Machine Learning and Knowledge
  Discovery in Databases}. Springer, 2023.

\bibitem[Watson et~al.(2023)Watson, Juergens, Bennett, Trippe, Yim, Eisenach,
  Ahern, Borst, Ragotte, Milles, Wicky, Hanikel, Pellock, Courbet, Sheffler,
  Wang, Bett, Bera, Roy, Savile, Xu, Dou, Eguchi, Powers, Ravichandran, Madden,
  Correia, Schief, Marquis, Chow, Brunette, DiMaio, Baek, and
  Baker]{watson2023rf}
James~L. Watson, David Juergens, Nathaniel Bennett, Brian Trippe, Jaekyung Yim,
  Helen Eisenach, William Ahern, Andrew Borst, Robert Ragotte, Lukas Milles,
  Basile I.~M. Wicky, Nikita Hanikel, Samuel~J. Pellock, Alexis Courbet,
  William Sheffler, Jianyi Wang, Keenan Bett, Asim Bera, Ambarish Roy,
  Christopher Savile, Yifan Xu, James Dou, Rebecca~R. Eguchi, Taylor Powers,
  Ranjani Ravichandran, Patrick Madden, Bruno Correia, William~R. Schief, David
  Marquis, Christine~M. Chow, T.~J. Brunette, Frank DiMaio, Minkyung Baek, and
  David Baker.
\newblock De novo design of protein structure and function with rfdiffusion.
\newblock \emph{Nature}, 620:\penalty0 687--696, 2023.
\newblock \doi{10.1038/s41586-023-06426-4}.

\bibitem[Xu et~al.(2022)Xu, Yu, Song, Shi, Ermon, and Tang]{xu2022geodiff}
Minkai Xu, Lantao Yu, Yang Song, Chence Shi, Stefano Ermon, and Jian Tang.
\newblock Geodiff: A geometric diffusion model for molecular conformation
  generation.
\newblock In \emph{International Conference on Learning Representations}, 2022.

\bibitem[Xu et~al.(2023)Xu, Powers, Dror, Ermon, and Leskovec]{xu2023geometric}
Minkai Xu, Alexander~S Powers, Ron~O Dror, Stefano Ermon, and Jure Leskovec.
\newblock Geometric latent diffusion models for 3d molecule generation.
\newblock In \emph{International Conference on Machine Learning}, 2023.

\bibitem[Yim et~al.(2023)Yim, Trippe, De~Bortoli, Mathieu, Doucet, Barzilay,
  and Jaakkola]{yim2023se}
Jason Yim, Brian~L Trippe, Valentin De~Bortoli, Emile Mathieu, Arnaud Doucet,
  Regina Barzilay, and Tommi Jaakkola.
\newblock Se (3) diffusion model with application to protein backbone
  generation.
\newblock In \emph{International Conference on Machine Learning}, 2023.

\bibitem[Zhdanov et~al.(2024)Zhdanov, Ruhe, Weiler, Lucic, Brandstetter, and
  Forr\'e]{zhdanov2024cliffordsteerable}
Maksim Zhdanov, David Ruhe, Maurice Weiler, Ana Lucic, Johannes Brandstetter,
  and Patrick Forr\'e.
\newblock {Clifford-Steerable Convolutional Neural Networks}.
\newblock In \emph{International Conference on Machine Learning}, 2024.

\end{thebibliography}
\bibliographystyle{iclr2025_conference}

\appendix

    


\end{document}